\documentclass[12pt]{article}
\usepackage[a4paper,margin=1in]{geometry}
\usepackage{graphicx}
\usepackage{booktabs}
\usepackage{amsmath}
\usepackage{authblk}
\usepackage{import}

\usepackage{float}

\usepackage{subcaption}

\usepackage[round]{natbib}
\usepackage[colorlinks=true, linkcolor=blue, citecolor=blue, urlcolor=blue]{hyperref}

\title{Game Reasoning Arena: A Framework and Benchmark for Assessing Reasoning Capabilities of Large Language Models via Game Play}
\author[1,2]{Lucia Cipolina‑Kun}
\author[1,2]{Marianna Nezhurina}
\author[1,2]{Jenia Jitsev}
\affil[1]{LAION}
\affil[2]{Juelich Supercomputing Center (JSC), Research Center Juelich (FZJ) }

\date{\today}

\begin{document}

\maketitle


The \emph{Game Reasoning Arena} library provides a framework for evaluating the decision making abilities of large language models (LLMs) through strategic board games implemented in Google’s OpenSpiel library. The framework enables systematic comparisons between LLM‑based agents and other agents (random, heuristic, reinforcement learning agents, etc.) in various game scenarios by wrapping multiple board and matrix games and supporting different agent types. It integrates API access to models via liteLLM, local model deployment via vLLM, and offers distributed execution through Ray. This paper summarises the library’s structure, key characteristics, and motivation of the repository, highlighting how it contributes to the empirical evaluation of the reasoning of LLM and game‑theoretic behaviour. The complete library documentation can be found at \url{https://game-reasoning-arena.readthedocs.io/} and the library is hosted at \url{https://github.com/SLAMPAI/game_reasoning_arena}. 

\section{Introduction}
Recent advances in large language models have spurred interest in evaluating their reasoning and planning abilities beyond standard natural language benchmarks.  Strategic games offer a controlled environment where agents must plan, adapt and anticipate opponents’ moves, making them a natural testbed for decision ma\emph{Game Reasoning Arena}addresses this need by providing a unified framework for playing and analysing games such as Tic–Tac–Toe, Connect Four, Kuhn Poker and matrix games like Prisoner’s Dilemma.  It uses Google‑DeepMind’s \emph{OpenSpiel} library as the underlying game engine.  OpenSpiel is an open‑source collection of environments and algorithms for reinforcement learning and search/planning in games; it supports single‑ and multi‑agent, zero‑sum and general‑sum games with perfect and imperfect information and includes tools to analyse learning dynamics.  By building on this engine, the framework exposes a flexible API to configure games, agents and evaluation settings.

The repository is organised into several modules:

\begin{itemize}
  \item \textbf{Game registry and environments.}  Games are registered via decorators in the loaders module, associating a name with an OpenSpiel game loader and a custom environment class; for example, Tic–Tac–Toe and Connect Four have dedicated environment classes.  The registry singleton handles dynamic loading and instantiation of games.
  \item \textbf{Agents and policies.}  Agents implement a \verb|compute_action| method and include random, human and LLM agents.  The policy manager assigns policies based on configuration and uses an agent registry to map strings to classes.  The LLM agent queries a language model using a unified backend and extracts an action and optional reasoning, while the random agent selects uniformly from legal actions.
  \item \textbf{Simulation logic.}  The simulation loop creates an environment, seeds the game, and then repeatedly queries each agent for an action.  It handles both turn‑based and 
  simultaneous games, applies actions to the environment, checks for illegal moves, updates rewards and logs outcomes.  Optional parallel execution allows multiple episodes or games to run concurrently on several CPU or GPU resources.
  \item \textbf{Architecture overview.}  An accompanying architecture document outlines how the registry, environment, agent and backend layers interconnect.  At a high level, environments derive from a common OpenSpiel wrapper, agents derive from a base agent class, and backends provide model‑agnostic inference.  These modules are orchestrated by simulation utilities that coordinate policy assignment and logging.
\end{itemize}
Overall, the repository emphasises modularity: researchers can add new games or agents by implementing a new environment or agent class and registering it via decorators. 

\section{Framework Design}\label{sec:framework}

The simulation framework builds on the multi-agent reinforcement-learning (RL) paradigm, which wraps the extensive game catalogue of \emph{OpenSpiel} in a Gymnasium-like interface. This section describes the game loop, environment design, agent interface and language-model inference back-ends, emphasising how these components support rigorous evaluation of LLM reasoning within strategic games.

\subsection{Game loop and environment}
The codebase leverages \emph{OpenSpiel} as the backend for the games emulator. The OpenSpiel API interface acts as the RL environment, where it encapsulates the game state, checks the validity of actions and mediates the transition dynamics. At each step, the OpenSpiel API's output is wrapped in a text prompt, this is passed to the LLM agent as the state observation. The text prompt includes the current state representation, the set of legal actions and for some games a summary of past moves. This design mirrors Gymnasium's API and supports both turn-based and simultaneous-move games.

\subsection{Agent interface and observation structure}
Agents implement a unified \texttt{compute\_action} method that receives the prompt as observation and returns an action as a dictionary with two items, the selected \textit{action} and the \textit{reasoning} behind the chosen action. The prompt asks the LLM to output its reasoning \textit{before} selecting an action, to avoid ex-post contamination. The environment returns reward signals calculated by OpenSpiel as dictionaries mapping each player to a scalar value, consistent with multi-agent RL practice. 

\subsection{Language model inference backends}
A key feature of our framework is support for multiple inference backends, enabling experiments with both hosted and locally deployed language models. The LiteLLM backend offers a unified API covering more than a hundred providers, including OpenAI, Anthropic, Google and Groq, and allows cost-effective and fast inference through automatic batching, caching and rate-limit handling. It also enables researchers to mix providers within a single experiment, facilitating controlled comparisons across models. The vLLM backend runs language models locally on a GPU via the vLLM inference engine, offering privacy, deterministic performance and offline evaluation capabilities, albeit requiring local model weights and sufficient GPU memory. By decoupling the agent logic from the underlying back-end, our design allows researchers to plug in any model that conforms to the same interface and to investigate how the choice of model and inference infrastructure influences game-play behaviour.

\section{Motivation}
Evaluating LLMs in strategic contexts is motivated by the desire to understand whether these models can perform complex planning and reasoning beyond text completion.  Board and card games involve sequential decision making, imperfect information and adversarial dynamics, making them a rich domain to test capabilities aligned with reinforcement learning and game theory.  \emph{Board Game Arena} provides several incentives for researchers:
\begin{itemize}
  \item \textbf{Multi‑agent testing.}  The library supports for LLMs vs random, LLM vs human, LLM vs LLM and self‑play settings, enabling comparative analyses of different models and baselines. 
  
  \item \textbf{Diverse games.}  By including both perfect information games (Tic-Tac-Toe, Connect Four) and hidden information games (Kuhn Poker), as well as matrix games (Prisoner’s Dilemma, Matching Pennies), the framework tests various aspects of strategic reasoning.
  
  \item \textbf{Flexible backends and cross‑provider evaluation.}  Models from Groq, Together AI, Fireworks and other providers can be used alongside local vLLM models; this allows researchers to assess how inference speed, token limit and temperature influence performance.
  
  \item \textbf{Scalability.}  Integration with Ray enables parallel simulation across multiple CPUs/GPUs or even SLURM clusters, facilitating large‑scale experiments.
  
  \item \textbf{Extensibility.}  The library is designed in a flexible and modular way  to add new games or agents.  This is essential for investigating novel game scenarios or implementing reinforcement learning agents for comparison.

   \item \textbf{Multi‑agent training.} The library is designed following Gymansium and RLLIB paradigms for reinforcement learning. This makes it suitable for users to train their own agent in the provided environment.
\end{itemize}

From a researcher's point of view, the framework bridges the gap between language modelling and game‑theoretic evaluation.  It allows testing whether LLMs can learn to cooperate in iterated Prisoner’s Dilemma, bluff in Kuhn Poker or plan winning sequences in Connect Four.  By logging reasoning strings from LLM responses and recording illegal moves, researchers can study failure modes and prompt designs.  Given the rapid development of open and proprietary models, the cross‑backend registry is a pragmatic feature that simplifies experimentation.  Overall, the \emph{Board Game Arena} repository offers a structured and extensible platform to interrogate the strategic capabilities of modern language models.


\section{Prompting System Architecture}
\label{sec:prompting}

A central feature of our benchmark is its structured prompting system, which enables consistent, analyzable interactions between language models and diverse board game environments. The design is modular and extensible, reflecting best practices in agent–environment interfaces and prompt engineering for LLMs.

\subsection{Observation and Decision Cycle}

Each game environment implements a state-to-observation transformation that produces a structured dictionary for each agent. This includes:

\begin{itemize}
  \item \texttt{state\_string}: A human-readable representation of the current game state.
  \item \texttt{legal\_actions}: A list of all valid moves the agent can take.
  \item \texttt{prompt}: A fully constructed prompt string containing task-specific context.
\end{itemize}

Agents extract the \texttt{prompt} field, which is passed to an LLM backend. Backend-specific adapters apply any required formatting (e.g., chat templates for dialogue models). The returned string is parsed into an action and accompanying reasoning using structured output constraints.

\subsection{Hierarchical Prompt Construction}

Prompt generation follows a hierarchical strategy, allowing base prompts to be reused while enabling customization for game-specific contexts.

\paragraph{Base Environment Prompts} The generic OpenSpiel environment provides prompts that include:
\begin{itemize}
  \item Game name and player role
  \item Move number and board visualization
  \item List of legal actions
\end{itemize}

\paragraph{Specialized Prompts} Environments for games with hidden information or contextual semantics—such as Kuhn Poker—extend the prompt with:
\begin{itemize}
  \item Private observations (e.g., the player's hidden card)
  \item Game history (e.g., betting sequence)
  \item Interpretable action labels based on state semantics
\end{itemize}

\subsection{Standardized Prompt Augmentation}

All prompts are further formatted using a standardized wrapper function that appends:
\begin{itemize}
  \item A reasoning directive, instructing the LLM to verbalize its strategy.
  \item A JSON schema for the output, specifying fields \texttt{reasoning} and \texttt{action}.
\end{itemize}

This dual-layer prompt format enables structured logging and consistent response parsing.

\subsection*{Example: Prompt in Kuhn Poker}

The following is an example of a prompt generated for a Kuhn Poker game. This prompt includes both private and public information, contextualized action labels, and structured response instructions:

\begin{quote}
\small
\begin{verbatim}
You are Player 0 in the game Kuhn Poker.
Your private card: Jack
This is move number: 2
Betting history: ['Check']
Total pot size: 2 chips
Your contribution: 1 chips

Available actions:
0: Check (stay in the game without betting)
1: Bet (add a chip to the pot)

What action do you choose? Reply only with '0' or '1'.

First, think through the game strategy and explain your reasoning.
Only after that, decide on the best action to take.

Reply only in the following JSON format:
{
  'reasoning': <str>,
  'action': <int>
}
\end{verbatim}
\end{quote}

This format enables the LLM to reason through game dynamics and ensures that the action can be extracted programmatically using standard JSON parsing or regular expressions.

\subsection{Backend Integration and Parsing}

The system supports both raw-text and chat-based models. When needed, prompts are automatically formatted into structured chat messages using templates tailored to each backend. To handle formatting inconsistencies, model outputs are parsed with regular expressions, allowing structured data to be extracted reliably. Any reasoning text generated by the model is preserved for further analysis and visualization.

\subsection{Extensibility for New Environments}

To support new games, developers can simply override the \texttt{\_generate\_prompt()} method in their environment. The framework handles downstream formatting and parsing, simplifying integration. The prompt system thus balances flexibility with enforced structure, allowing scalable and interpretable LLM-agent integration across games.

\section{Evaluation and Experimental Design}\label{sec:evaluation}

Our evaluation focuses on measuring the strategic competence and reasoning quality of language models beyond simple win rates. For each game we record per-step rewards, outcomes and the sequence of actions and reasoning strings produced by the agents. From these trajectories we derive metrics such as (i) average and maximum cumulative reward, (ii) decision optimality, measured as the proportion of moves matching the equilibrium or optimal policy, (iii) reasoning length and coherence, captured via simple heuristics on the textual rationales, and (iv) error rates, counting illegal or suboptimal moves.

To ensure statistical significance, we run multiple simulations per configuration and report means and standard errors. When comparing models across games or back-ends, we use paired-sample tests and bootstrap confidence intervals to assess differences in performance.

\subsection{Experimental design}
The framework supports systematic experimentation across a wide range of configurations, enabling controlled evaluation of agent behaviors under varying game setups. It facilitates large-scale simulations, hyperparameter sweeps, and ablation studies through seamless parallelization. Each experiment is initialized with a consistent random seed to ensure reproducibility, and simulations are executed using the framework introduced in Section~\ref{sec:framework}. To enable scalability, we leverage parallelization via Ray, distributing simulation workloads across available hardware resources. On high-performance computing clusters, experiments are scheduled using SLURM to facilitate efficient large-scale evaluations.

\subsection{Analysis tools}
Beyond aggregate metrics, we provide tools to inspect agent behaviour. Reasoning strings from the LLM agents are logged alongside actions, allowing qualitative analysis of decision-making and failure modes. Post-game processing scripts summarise common reasoning categories, identify when models refer to game-specific concepts and flag hallucinations or rule violations. These tools facilitate deeper insights into model capabilities and guide prompt engineering and model selection.

\subsection{Results}

To analyse the reasoning processes of LLMs in board game environments, we 
propose a categorization framework that labels model-generated justifications 
according to distinct reasoning patterns as shown in 
Table~\ref{tab:reasoning_types}. These include \emph{positional reasoning}, 
which emphasizes spatial control of the board (e.g., references to the 
"center column" or "corner"); \emph{opponent modeling}, where the model 
anticipates or reacts to the opponent's strategy; and \emph{blocking}, which 
denotes defensive actions aimed at preventing the opponent's success. 
Additional categories include \emph{winning logic}, which refers to direct 
strategies for achieving victory (e.g., "winning move", "fork"); 
\emph{heuristic-based reasoning}, characterized by general evaluative 
language (e.g., "best move", "advantageous"); and \emph{rule-based reasoning}, 
which appeals to known strategies or game rules. Finally, we include a 
\emph{random/unjustified} category for responses lacking clear strategic 
motivation. By associating these types of reasoning with specific lexical 
cues, this classification enables a structured analysis of LLM decision 
making, highlighting patterns, strengths, and potential shortcomings in model 
reasoning across varied game scenarios.

We apply this classification framework to large-scale gameplay logs from 
multiple LLMs and games, producing the following multi-perspective analysis.

\begin{table}[ht]
\centering
\small
\caption{Reasoning types with descriptions and representative words.}
\label{tab:reasoning_types}
\begin{tabular}{p{2.8cm} p{5.0cm} p{5.2cm}}
\toprule
\textbf{Reasoning Type} & \textbf{Description} & \textbf{Words} \\ \midrule
Positional & Focuses on spatial board control (e.g.\ “center column”) & center column, “center square”, “corner”, “edge” \\ \midrule
Opponent modeling & Refers to or anticipates opponent strategy & “opponent”, “they are trying”, “their strategy”, “their move” \\ \midrule
Blocking & Refers to countering opponent threats & “block”, “prevent”, “stop opponent”, “avoid opponent”, “counter” \\ \midrule
Winning logic & Direct mentions of creating win conditions & “win”, “winning move”, “connect”, “fork”, “threat”, “chance of winning” \\ \midrule
Heuristic-based & General heuristics like “center is best” & “best move”, “most likely”, “advantageous”, “better chance” \\ \midrule
Rule-based & Based on known strategies (forks, chains) & according to, “rule”, “strategy” \\ \midrule
Random/unjustified & Reasoning lacks strategic justification & “random”, “guess” \\ \bottomrule
\end{tabular}
\end{table}


\subsubsection{Temporal Adaptation Across Games (Radar Overview)}

\begin{figure}[H]
  \centering
  \includegraphics[width=0.85\textwidth]
  {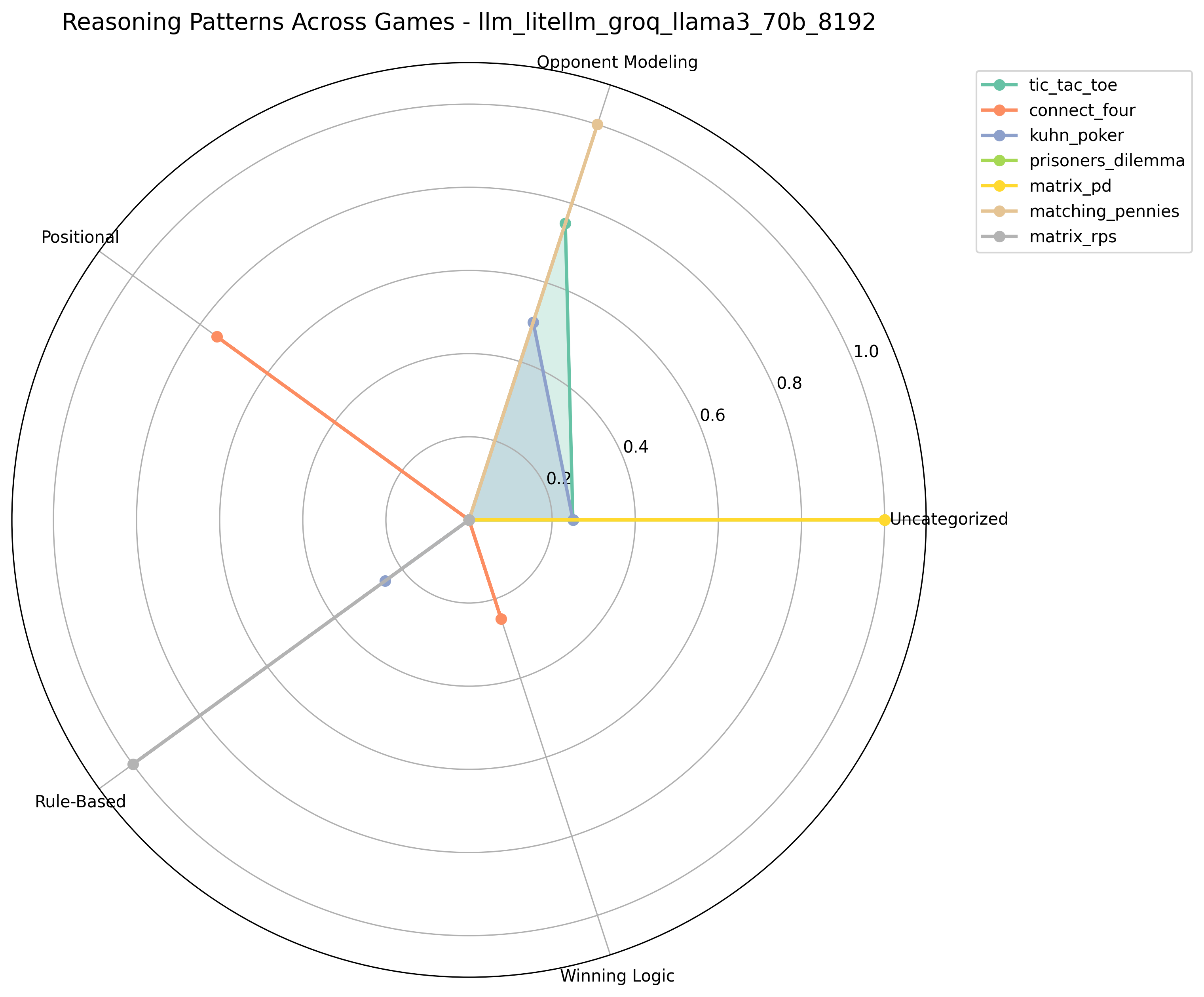}
  \caption{Temporal reasoning-type patterns across games for
  \texttt{groq\_llama3\_70b\_8192}.}
  \label{fig:radar_llm_litellm_groq_llama3_70b_8192}
\end{figure}

Figure~\ref{fig:radar_llm_litellm_groq_llama3_70b_8192} summarises 
\emph{temporal} reasoning patterns for 
\texttt{groq\_llama3\_70b\_8192} across seven games. Each 
polygon corresponds to a game; axes are reasoning types and radii denote 
normalised frequencies over the course of play. We observe opponent-focused 
behaviour in \texttt{tic\_tac\_toe} and \texttt{prisoners\_dilemma}, 
positional emphasis in \texttt{connect\_four}, rule-based regularity in 
\texttt{matching\_pennies}, and a larger uncategorised share in 
\texttt{matrix\_pd}.

\subsubsection{Aggregate Per-Game Profiles (Stacked Bars and By-Game Views)}

\begin{figure}[H]
  \centering
  \includegraphics[width=0.85\textwidth]
  {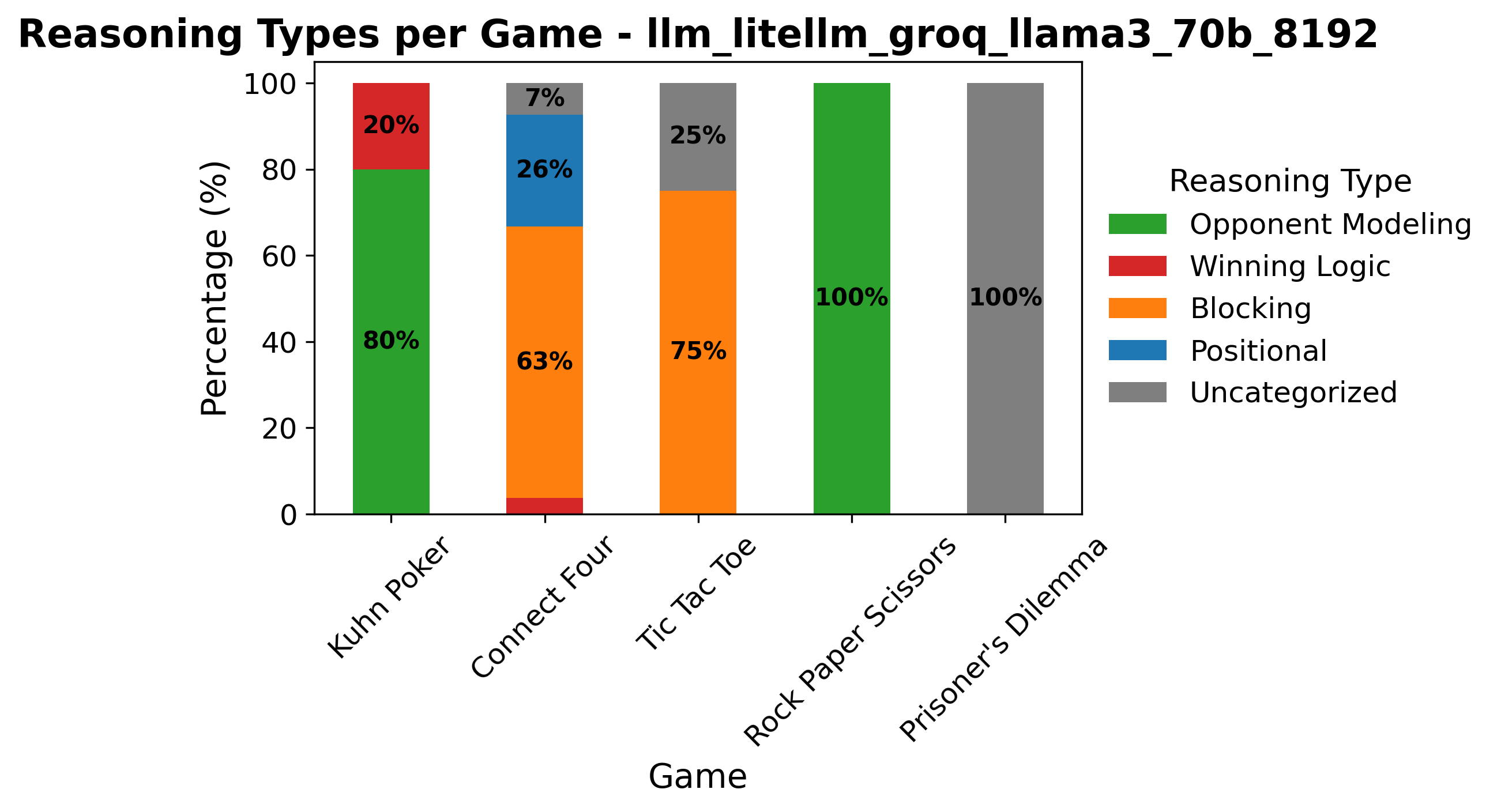}
  \caption{Stacked reasoning percentages by game:
  \texttt{groq\_llama3\_70b\_8192}.}
  \label{fig:stacked-70b}
\end{figure}

\begin{figure}[H]
  \centering
  \includegraphics[width=0.85\textwidth]
  {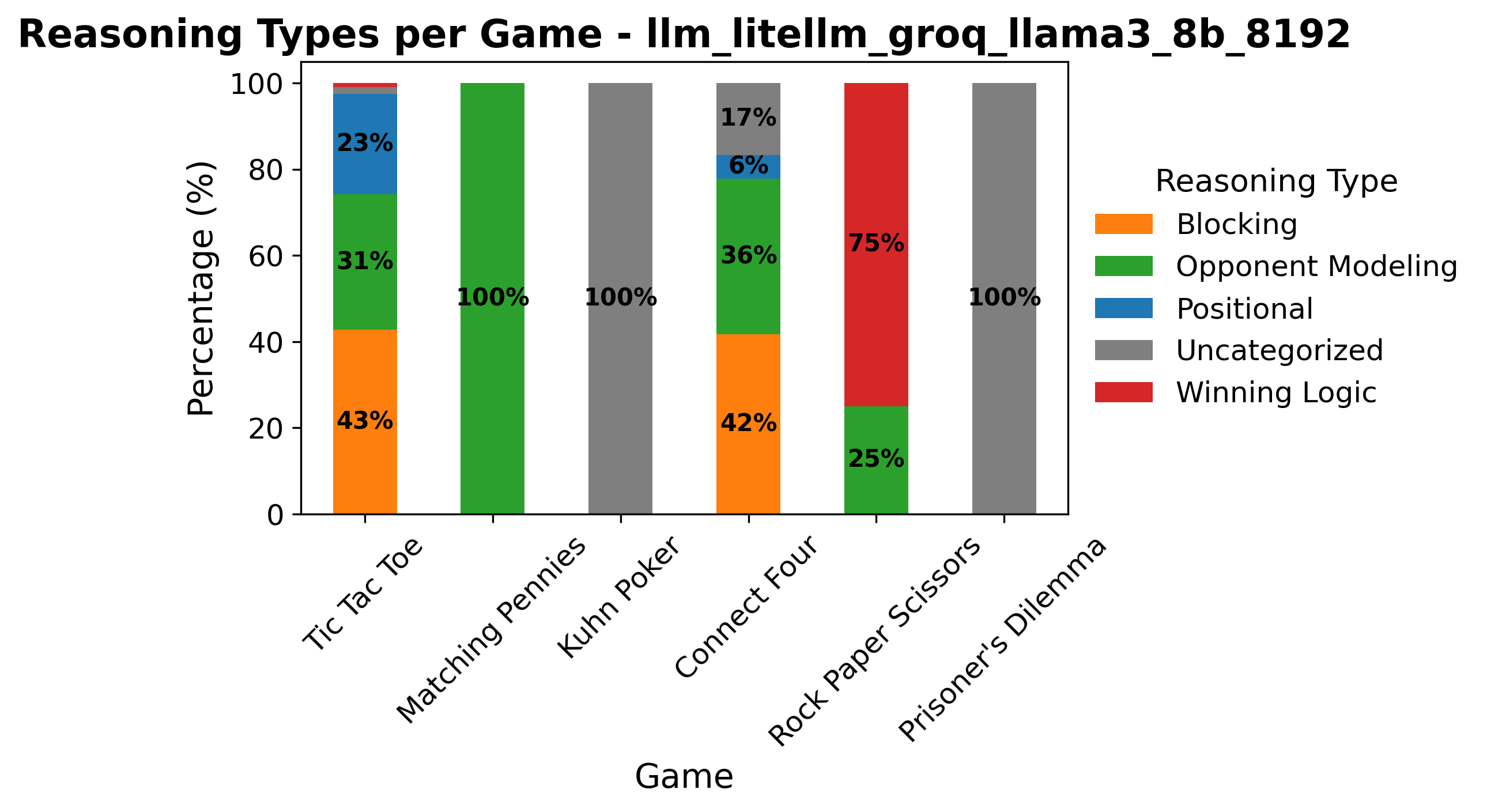}
  \caption{Stacked reasoning percentages by game:
  \texttt{groq\_llama3\_8b\_8192}.}
  \label{fig:stacked-8b}
\end{figure}

Figures~\ref{fig:stacked-70b}--\ref{fig:stacked-8b} report the overall 
per-game distribution of reasoning types. Stacked bars offer a compact 
percentage view for each game; the by-game plots show the same information 
with separate bars per category.

The 70B model exhibits richer variety and stronger opponent modelling in 
\texttt{tic\_tac\_toe} and \texttt{matching\_pennies}, positional play in 
\texttt{connect\_four}, and uncategorised traces in the matrix games. The 
8B model shows heavier reliance on few categories and more frequent 
uncategorised segments.

\begin{figure}[H]
    \centering
    \includegraphics[width=1\textwidth]{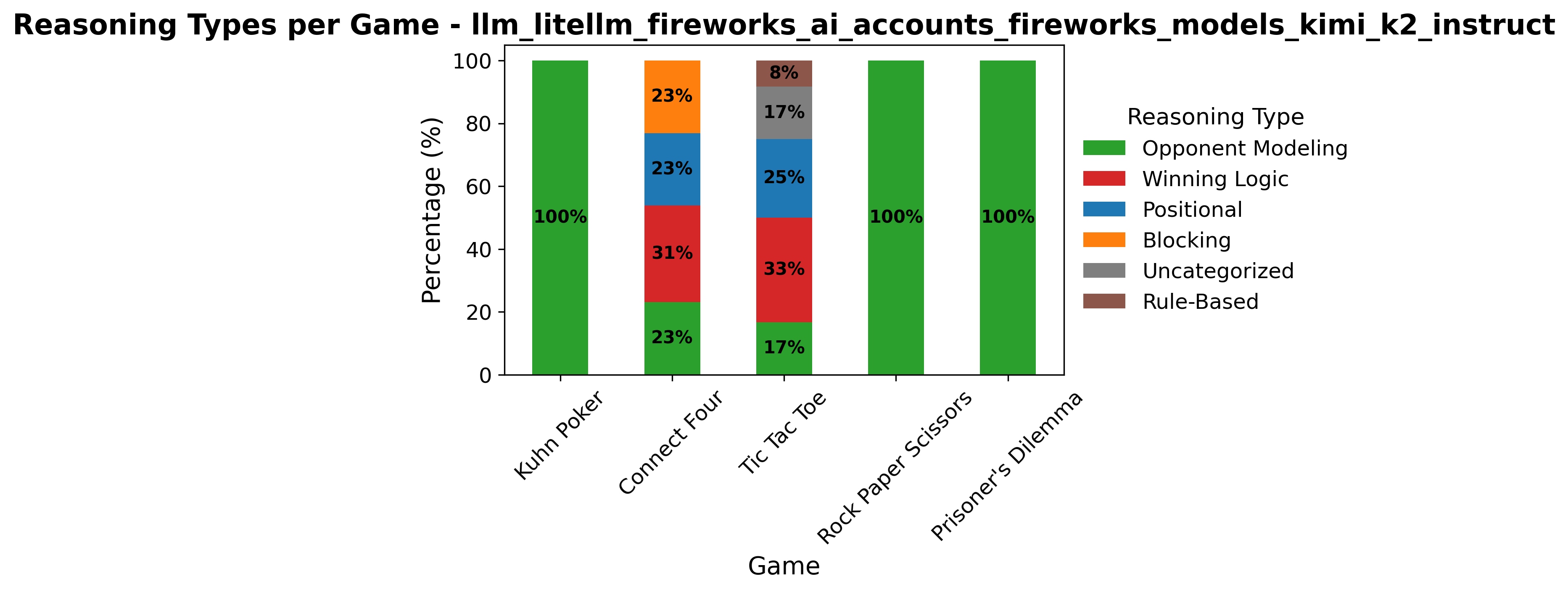}
    \caption{Reasoning distribution across games for 
    \texttt{kimi\_k2\_instruct}. 
    Opponent modeling dominates in simpler games, while more complex board 
    games elicit a broader range of reasoning types.}
    \label{fig:reasoning_stacked_kimi}
\end{figure}

\begin{figure}[H]
    \centering
    \includegraphics[width=0.8\textwidth]{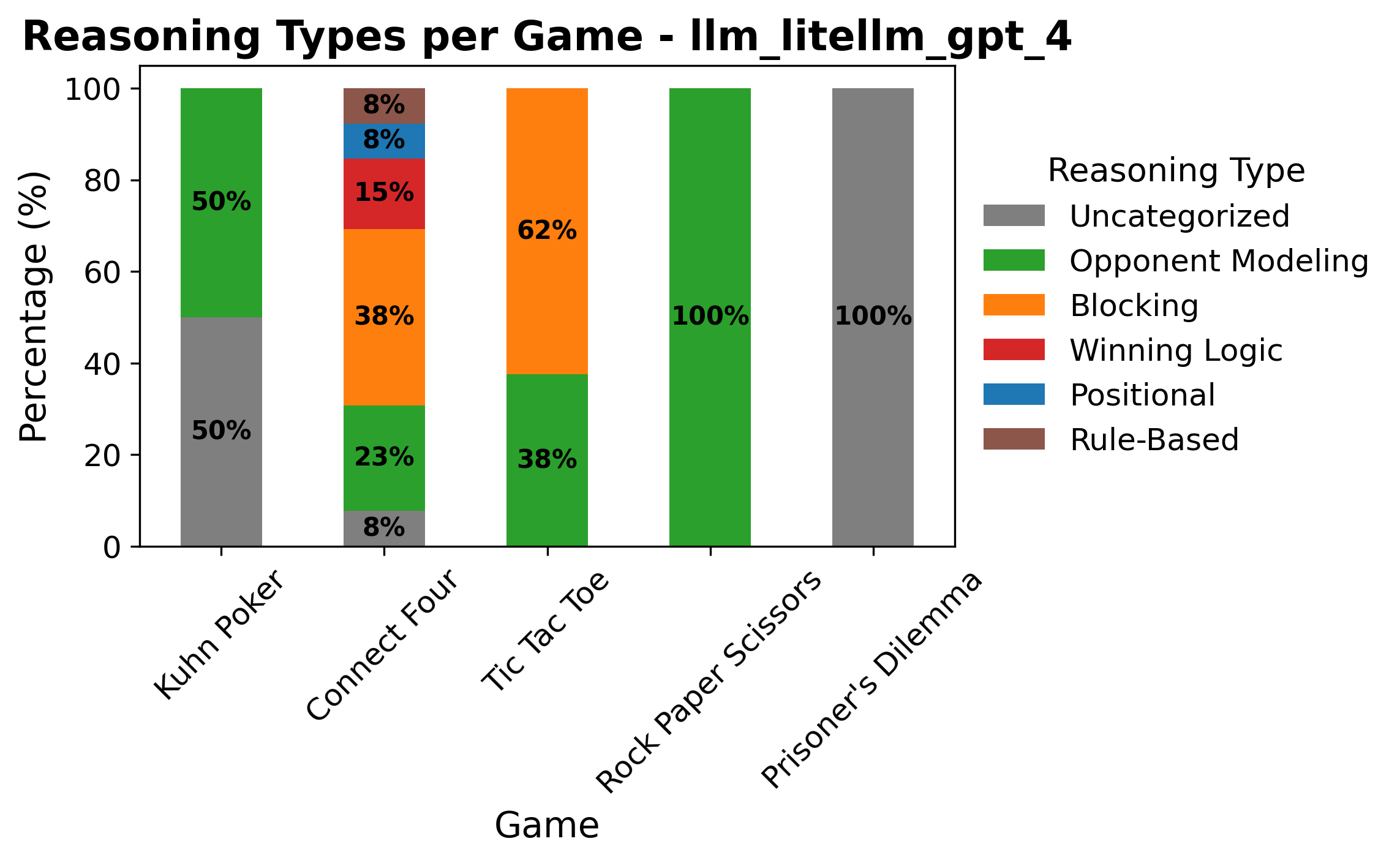}
    \caption{Reasoning distribution for \texttt{gpt\_4}. 
    The model adapts its reasoning to game structure, mixing opponent 
    modeling, blocking, and uncategorized strategies.}
    \label{fig:reasoning_stacked_gpt4}
\end{figure}

\begin{figure}[H]
    \centering
    \includegraphics[width=0.85\textwidth]{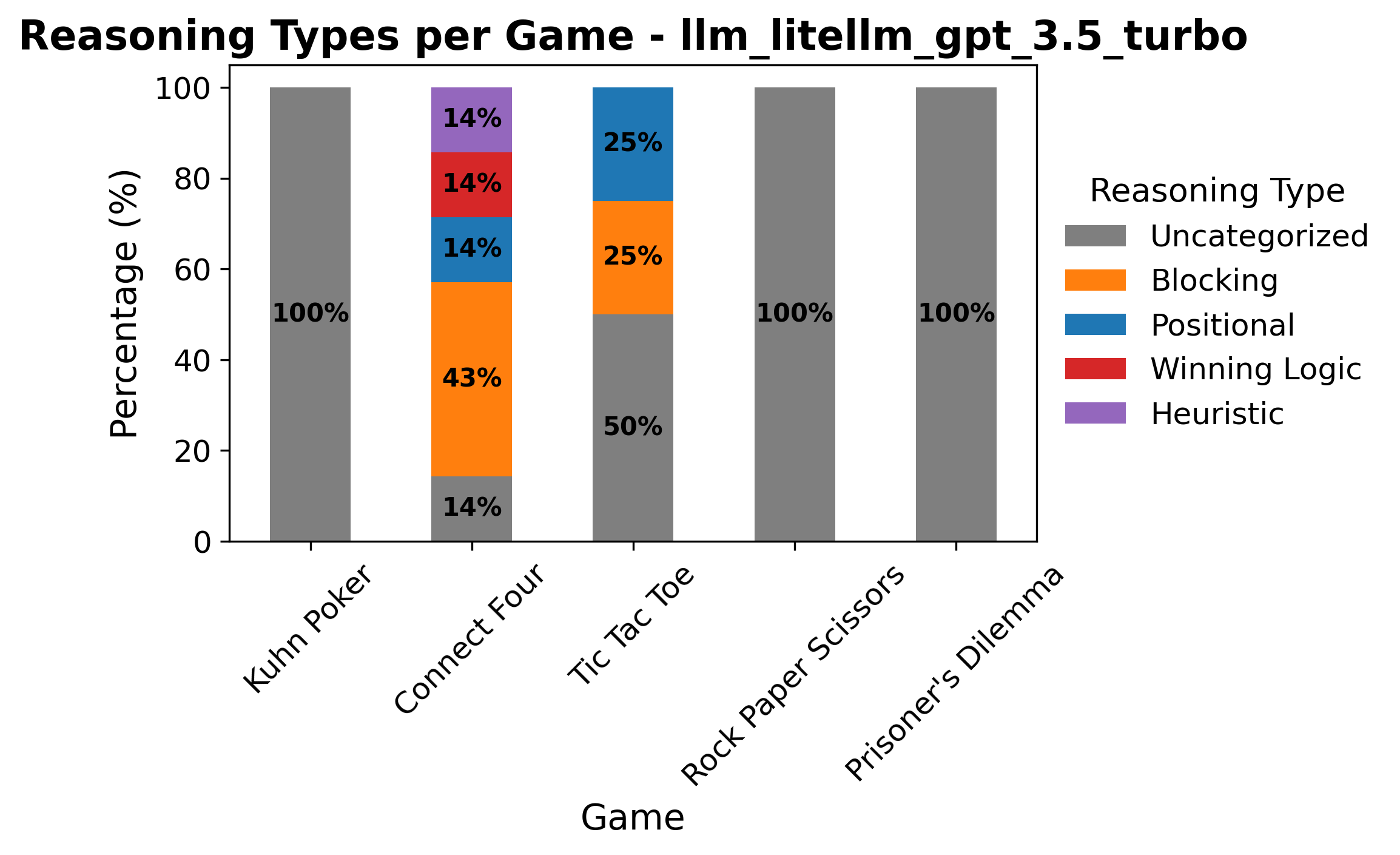}
    \caption{Reasoning distribution for \texttt{gpt\_3.5\_turbo}. 
    Compared to GPT-4, the model shows weaker adaptation and a higher 
    proportion of uncategorized reasoning.}
    \label{fig:reasoning_stacked_gpt35}
\end{figure}

Figures~\ref{fig:reasoning_stacked_kimi}, 
\ref{fig:reasoning_stacked_gpt4}, and \ref{fig:reasoning_stacked_gpt35} show 
that stronger models adapt their reasoning to the structural demands of each 
game. For example, the \texttt{Kimi-K2-Instruct} model relies almost 
exclusively on opponent modeling in Kuhn Poker and Prisoner’s Dilemma, while 
exhibiting more diverse strategies in Connect Four and Tic-Tac-Toe. \texttt{GPT-4}, in 
turn, displays greater adaptability by deploying blocking strategies in Connect 
Four while relying on opponent modeling in sequential games. \texttt{GPT-3.5-Turbo}, 
however, exhibits weaker specialization, with many outputs uncategorized and 
lacking clear strategic grounding. These findings suggest that stronger models 
not only adapt within games but also calibrate their strategies across games.

\subsubsection{Within-Game Evolution (Same Game, Different Turns)}

We bin turns and compute per-bin reasoning proportions to detect shifts in 
strategy within a single game. Heatmaps emphasise concentration and 
switching; stacked bars show the mixture per bin. In \texttt{tic\_tac\_toe}, 
both models shift towards pure \emph{Blocking} in the endgame, but their 
openings differ—one starts with \emph{Winning Logic} then \emph{Opponent 
Modeling}, while the other begins with strong \emph{Opponent Modeling} and 
a positional midgame.

\begin{figure}[H]
  \centering
  \includegraphics[width=\textwidth]
  {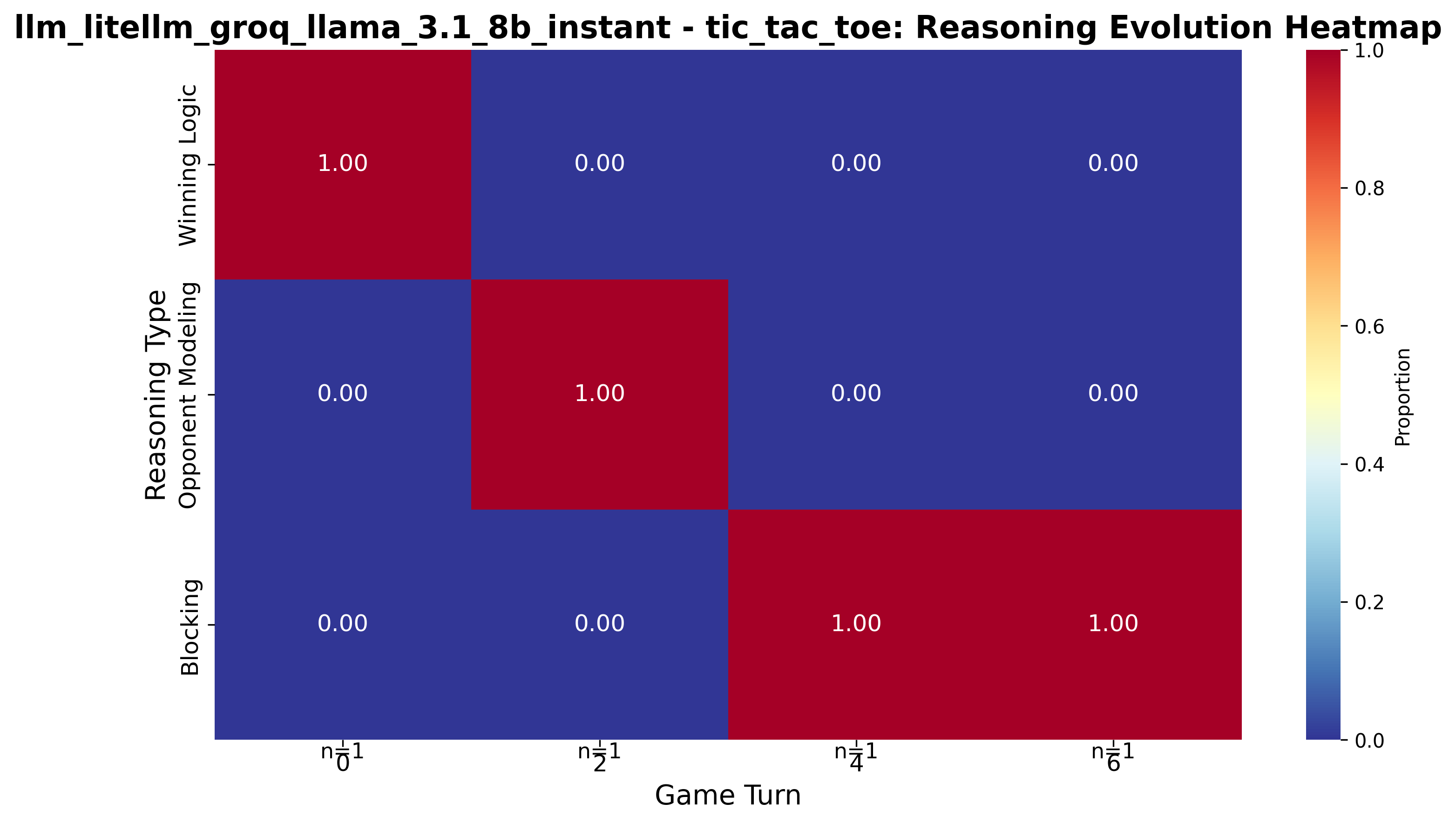}
  \caption{Within-game reasoning heatmap (tic\_tac\_toe):
  \texttt{groq\_llama\_3\_1\_8b\_instant}.}
  \label{fig:heat_instant_ttt}
\end{figure}

\begin{figure}[H]
  \centering
  \includegraphics[width=\textwidth]
  {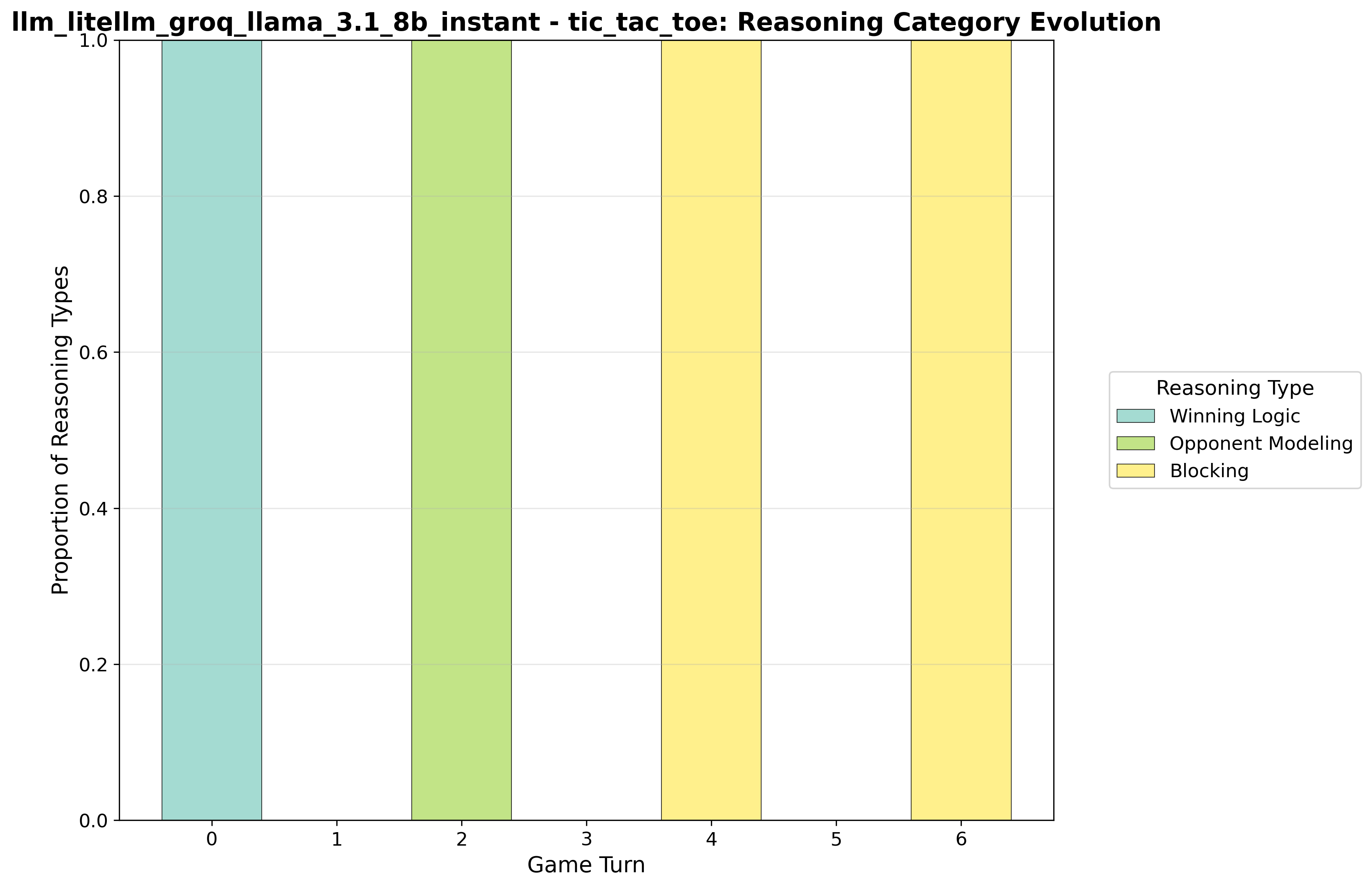}
  \caption{Turn-binned proportions (tic\_tac\_toe):
  \texttt{groq\_llama\_3\_1\_8b\_instant}.}
  \label{fig:bars_instant_ttt}
\end{figure}

\begin{figure}[H]
  \centering
  \includegraphics[width=\textwidth]
  {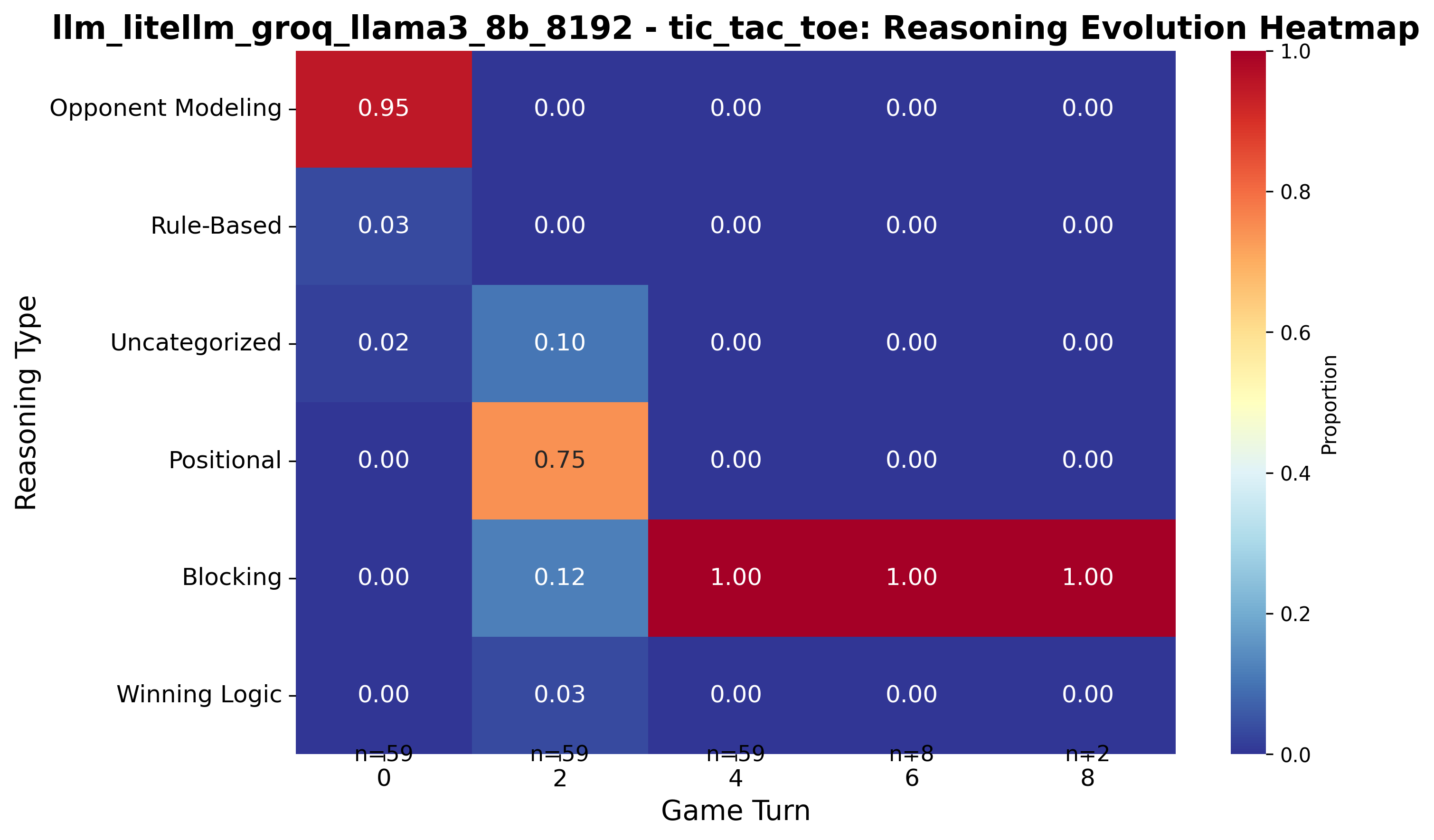}
  \caption{Within-game reasoning heatmap (tic\_tac\_toe):
  \texttt{groq\_llama3\_8b\_8192}.}
  \label{fig:heat_8b8192_ttt}
\end{figure}

\begin{figure}[H]
  \centering
  \includegraphics[width=\textwidth]
  {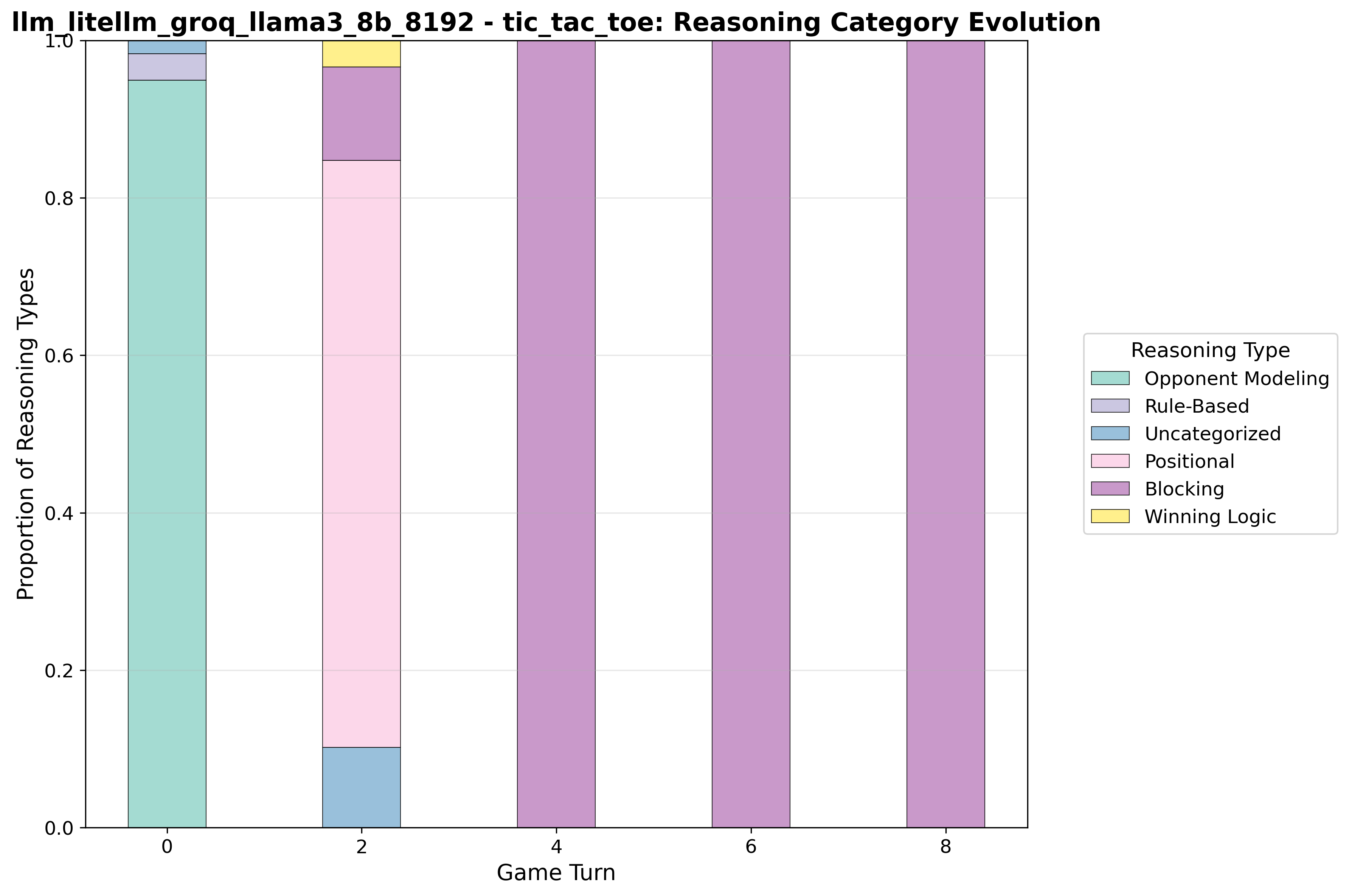}
  \caption{Turn-binned proportions (tic\_tac\_toe):
  \texttt{groq\_llama3\_8b\_8192}.}
  \label{fig:bars_8b8192_ttt}
\end{figure}

\subsubsection{Per-Game Pie Charts (Detailed Breakdown Examples)}

We include game-specific pies to illustrate the detailed mix behind the 
aggregates. In \texttt{tic\_tac\_toe} the instant 8B model balances 
\emph{Blocking} (50\%), \emph{Winning Logic} (25\%), and 
\emph{Opponent Modeling} (25\%). The 8B~8192 variant diversifies further, 
and in \texttt{kuhn\_poker} the 70B model emphasises 
\emph{Opponent Modeling} with smaller rule-based and uncategorised shares.

\begin{figure}[H]
  \centering
  \includegraphics[width=0.65\linewidth]
  {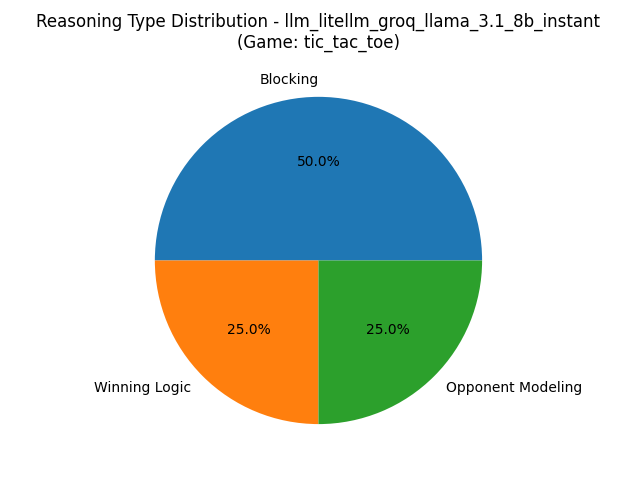}
  \caption{Pie breakdown (tic\_tac\_toe):
  \texttt{groq\_llama\_3.1\_8b\_instant}.}
  \label{fig:reasoning-tic-instant}
\end{figure}

\begin{figure}[H]
  \centering
  \includegraphics[width=0.65\linewidth]
  {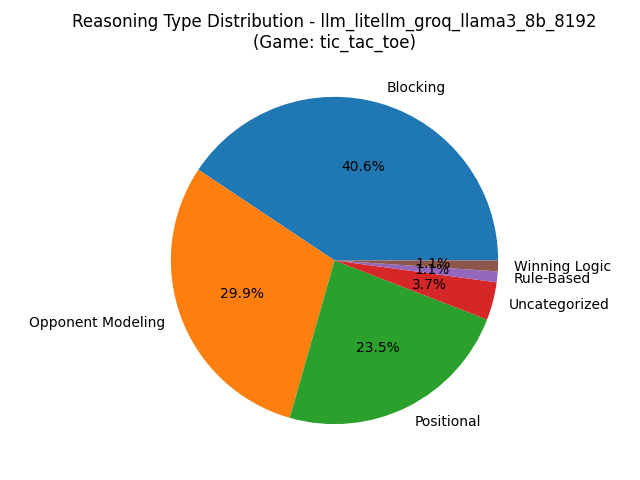}
  \caption{Pie breakdown (tic\_tac\_toe):
  \texttt{groq\_llama3\_8b\_8192}.}
  \label{fig:reasoning-tic-8b8192}
\end{figure}

\begin{figure}[H]
  \centering
  \includegraphics[width=0.65\linewidth]
  {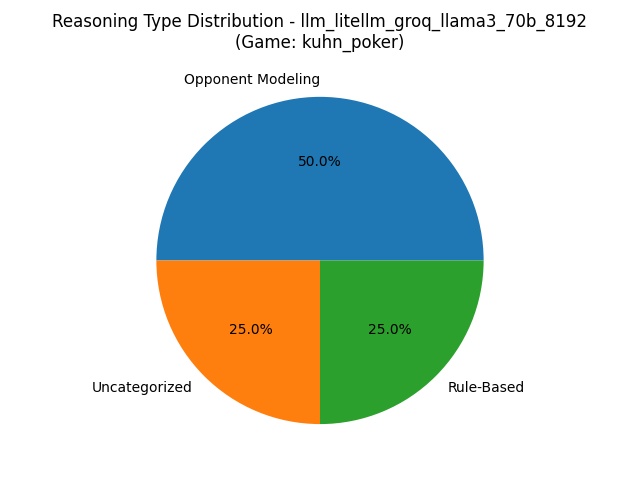}
  \caption{Pie breakdown (kuhn\_poker):
  \texttt{groq\_llama3\_70b\_8192}.}
  \label{fig:reasoning-kuhn-70b}
\end{figure}

\subsubsection{Entropy Across Agents Within a Game}

To quantify cross-model \emph{diversity of reasoning over time} for a fixed 
game, we compute the Shannon entropy of the reasoning-type distribution at 
each turn bin, aggregating across agents. Higher entropy indicates a larger 
spread across reasoning types, whereas zero entropy reflects unanimity.

\begin{figure}[H]
  \centering
  \includegraphics[width=0.75\textwidth]
  {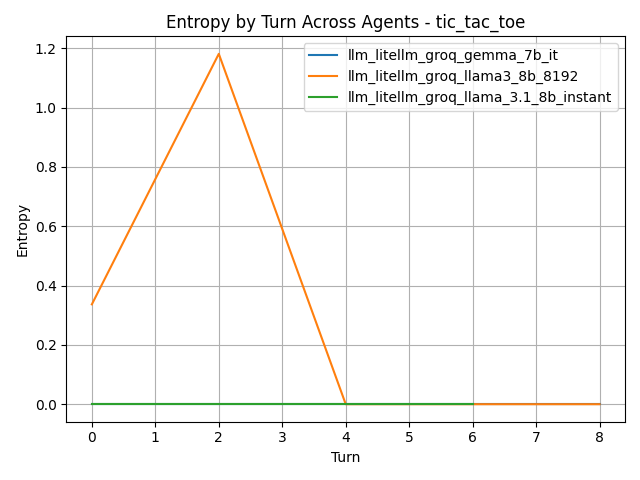}
  \caption{Entropy by turn across agents (tic\_tac\_toe). Higher values
  indicate more diverse reasoning distributions across agents at a given
  turn; zero denotes unanimity.}
  \label{fig:entropy_all_agents_ttt}
\end{figure}

Figure~\ref{fig:entropy_all_agents_ttt} reports entropy across agents for 
\texttt{tic\_tac\_toe}. The 
\texttt{groq\_llama3\_8b\_8192} model shows a clear mid-game 
peak (greater strategic diversity), followed by a collapse to near-zero in 
late turns, consistent with convergence to \emph{Blocking}. The other 
agents remain near zero throughout.

\subsubsection{Synthesis}

Taken together, these views reveal a coherent pattern. First, LLMs adapt 
their reasoning \emph{across games} in ways aligned with game structure 
(radar, stacked bars). Second, they adapt \emph{within a single game}: 
openings display planning/anticipation, mid-games emphasise board control, 
and endgames converge to defensive blocking (heatmaps and turn-stacked 
bars). Third, model size influences diversity—larger models tend to display 
richer, more balanced profiles, while smaller ones rely on a narrower set 
of rationales. Finally, entropy confirms that strategic diversity varies 
over time and across agents, peaking when the search space is most open and 
collapsing as terminal constraints dominate.


\newpage

\section{Conclusion and Outlook}\label{sec:conclusion}

We have introduced a board-game codebase and benchmark for evaluating the strategic reasoning of large language models. Building on the multi-agent RL paradigm of OpenSpiel, our framework unifies a diverse set of discrete games within a common interface, supports multiple LLM inference backends and records both actions and reasoning strings. The resulting benchmark enables systematic comparisons across models, games and inference modalities, providing insights into how language models plan, adapt and cooperate in strategic settings.

Our experiments outline a broad suite of evaluation metrics and demonstrate how distributed execution facilitates large-scale benchmarking. Although our focus has been on turn-based board and card games, the modular design readily extends to other game genres and to new agent types. Future work will explore richer reasoning analyses, adaptive prompting strategies and integration with reinforcement-learning training loops. We hope that this benchmark will stimulate further research into the game-theoretic capabilities of language models and their applications in human-AI interaction.


\newpage
\section{Related Research}

The growing interest in using games to evaluate large language models has led to several complementary benchmarks.  

\textbf{TextArena} ~\cite{GuertlerEtAl2025TextArena} is an open‑source collection of more than fifty text‑based games designed to train and evaluate agentic behaviour in LLMs. It supports single‑player, two‑player and multi‑player scenarios and provides an online platform where models and humans can compete, with performance tracked via a real‑time TrueSkill leaderboard.

\textbf{GameArena}~\cite{HuEtAl2025GameArena} focuses on reasoning skills. It is a dynamic benchmark in which models play live computer games against human opponents to test deductive and inductive reasoning. The benchmark comprises three games, collects over 2{,}000 gameplay sessions, and analyses the resulting data to measure fine‑grained reasoning capabilities.

In the realm of board and grid‑based games, \cite{topsakal2024evaluatinglargelanguagemodels} introduced an extensible benchmark using games such as Tic–Tac–Toe, Connect Four and Gomoku to compare LLMs. The open‑source simulator runs thousands of matches between models from Anthropic, Google, OpenAI and Meta, generating results in multiple formats and creating a leaderboard for analysis.

\textbf{GameBench}~\cite{GameBench2024} extends evaluation to a broader set of strategic games. It defines a cross‑domain benchmark covering nine strategic board, card and social deduction games that test reasoning along axes such as imperfect information, stochasticity and communication. Evaluations of GPT‑3 and GPT‑4 (with chain‑of‑thought or planning scaffolds) reveal that even the best models often underperform random baselines and fail to approach human-level performance.

\textbf{Board Game Bench}~\cite{BoardGameBench2025} is an a
rena‑based benchmark that pits LLMs head‑to‑head in board games. Despite rapid progress, LLMs remain poor at most board games; success requires rule comprehension, opponent modelling, probability estimation, and long‑term planning. The system records wins/losses and uses a Bayesian Elo‑like rating to rank models.

\textbf{lmgame‑Bench}~\cite{lmgameBench2025} converts real video games into robust evaluation benchmarks. The framework integrates perception, memory and reasoning scaffolds and provides a unified environment across games such as \textit{Super Mario Bros.}, \textit{Tetris}, and \textit{Ace Attorney}. Reinforcement‑learning experiments on this suite demonstrate both in‑domain and cross‑domain generalisation gains.

\medskip

\begin{table}[t]
\centering
\small
\begin{tabular}{p{3cm} p{2.5cm} p{5.5cm} p{3.5cm}}
\toprule
\textbf{Benchmark} & \textbf{Game Set} & \textbf{Focus/Skills Tested} & \textbf{Citation} \\
\midrule
\textbf{TextArena} & 50+ text games & Social reasoning (negotiation, deception, theory of mind); real-time leaderboard & \cite{GuertlerEtAl2025TextArena} \\
\textbf{GameArena} & 3 live games, 2{,}000+ sessions & Deductive and inductive reasoning; interactive gameplay with human opponents & \cite{HuEtAl2025GameArena} \\
\textbf{Grid-based LLM benchmark} & Tic–Tac–Toe, Connect Four, Gomoku & Rule understanding, win-rate analysis, strategic variation by prompt & \cite{topsakal2024evaluatinglargelanguagemodels} \\
\textbf{GameBench} & 9 board/card/social games & Reasoning under imperfect info, stochasticity, communication; evaluation with GPT-3/4 & \cite{GameBench2024} \\
\textbf{Board Game Bench} & Head-to-head board games & Opponent modelling, long-term planning, Bayesian Elo ranking & \cite{BoardGameBench2025} \\
\textbf{lmgame‑Bench} & Real video games & Vision, memory, planning, and generalisation under contamination-robust setup & \cite{lmgameBench2025} \\
\bottomrule
\end{tabular}
\caption{Comparison of recent game-based LLM evaluation benchmarks.}
\label{tab:game_benchmarks}
\end{table}

Collectively, these benchmarks indicate that while LLMs excel at many language tasks, strategic gameplay remains challenging. They highlight the importance of dynamic, multi‑agent and interactive environments for evaluating reasoning, planning and social skills, and suggest motivation for frameworks such as \emph{Game Reasoning Arena}


\newpage
\bibliographystyle{plainnat}  
\bibliography{main}

\begin{thebibliography}{6}
\providecommand{\natexlab}[1]{#1}
\providecommand{\url}[1]{\texttt{#1}}
\expandafter\ifx\csname urlstyle\endcsname\relax
  \providecommand{\doi}[1]{doi: #1}\else
  \providecommand{\doi}{doi: \begingroup \urlstyle{rm}\Url}\fi

\bibitem[{Board Game Bench authors}(2025)]{BoardGameBench2025}
{Board Game Bench authors}.
\newblock {Board Game Bench}: What is board game bench? and how it works, 2025.
\newblock https://www.boardgamebench.com/.

\bibitem[Costarelli et~al.(2024)Costarelli, Allen, Hauksson, Sodunke, Hariharan, Cheng, Li, Clymer, and Yadav]{GameBench2024}
Anthony Costarelli, Mat Allen, Roman Hauksson, Grace Sodunke, Suhas Hariharan, Carlson Cheng, Wenjie Li, Joshua Clymer, and Arjun Yadav.
\newblock Gamebench: Evaluating strategic reasoning abilities of llm agents, 2024.
\newblock URL \url{https://arxiv.org/abs/2406.06613}.

\bibitem[Guertler et~al.(2025)Guertler, Cheng, Yu, Liu, Choshen, and Tan]{GuertlerEtAl2025TextArena}
Leon Guertler, Bobby Cheng, Simon Yu, Bo~Liu, Leshem Choshen, and Cheston Tan.
\newblock {TextArena}: Competitive text‑based games for evaluating agentic behavior in llms.
\newblock \emph{arXiv}, 2025.

\bibitem[Hu et~al.(2025{\natexlab{a}})Hu, Huo, Zhang, Yu, Xing, Stoica, Rosing, Jin, and Zhang]{lmgameBench2025}
Lanxiang Hu, Mingjia Huo, Yuxuan Zhang, Haoyang Yu, Eric~P. Xing, Ion Stoica, Tajana Rosing, Haojian Jin, and Hao Zhang.
\newblock lmgame-bench: How good are llms at playing games?, 2025{\natexlab{a}}.
\newblock URL \url{https://arxiv.org/abs/2505.15146}.

\bibitem[Hu et~al.(2025{\natexlab{b}})Hu, Li, Xie, Jiang, Stoica, Jin, and Zhang]{HuEtAl2025GameArena}
Lanxiang Hu, Qiyu Li, Anze Xie, Nan Jiang, Ion Stoica, Haojian Jin, and Hao Zhang.
\newblock {GameArena}: Evaluating llm reasoning through live computer games.
\newblock In \emph{ICLR 2025}, 2025{\natexlab{b}}.
\newblock arXiv preprint arXiv:2412.06394.

\bibitem[Topsakal et~al.(2024)Topsakal, Edell, and Harper]{topsakal2024evaluatinglargelanguagemodels}
Oguzhan Topsakal, Colby~Jacob Edell, and Jackson~Bailey Harper.
\newblock Evaluating large language models with grid-based game competitions: An extensible llm benchmark and leaderboard, 2024.
\newblock URL \url{https://arxiv.org/abs/2407.07796}.

\end{thebibliography}


\newpage
\appendix

\section{Appendix}

\begin{figure}[ht]
  \centering
  \includegraphics[width=0.85\textwidth]{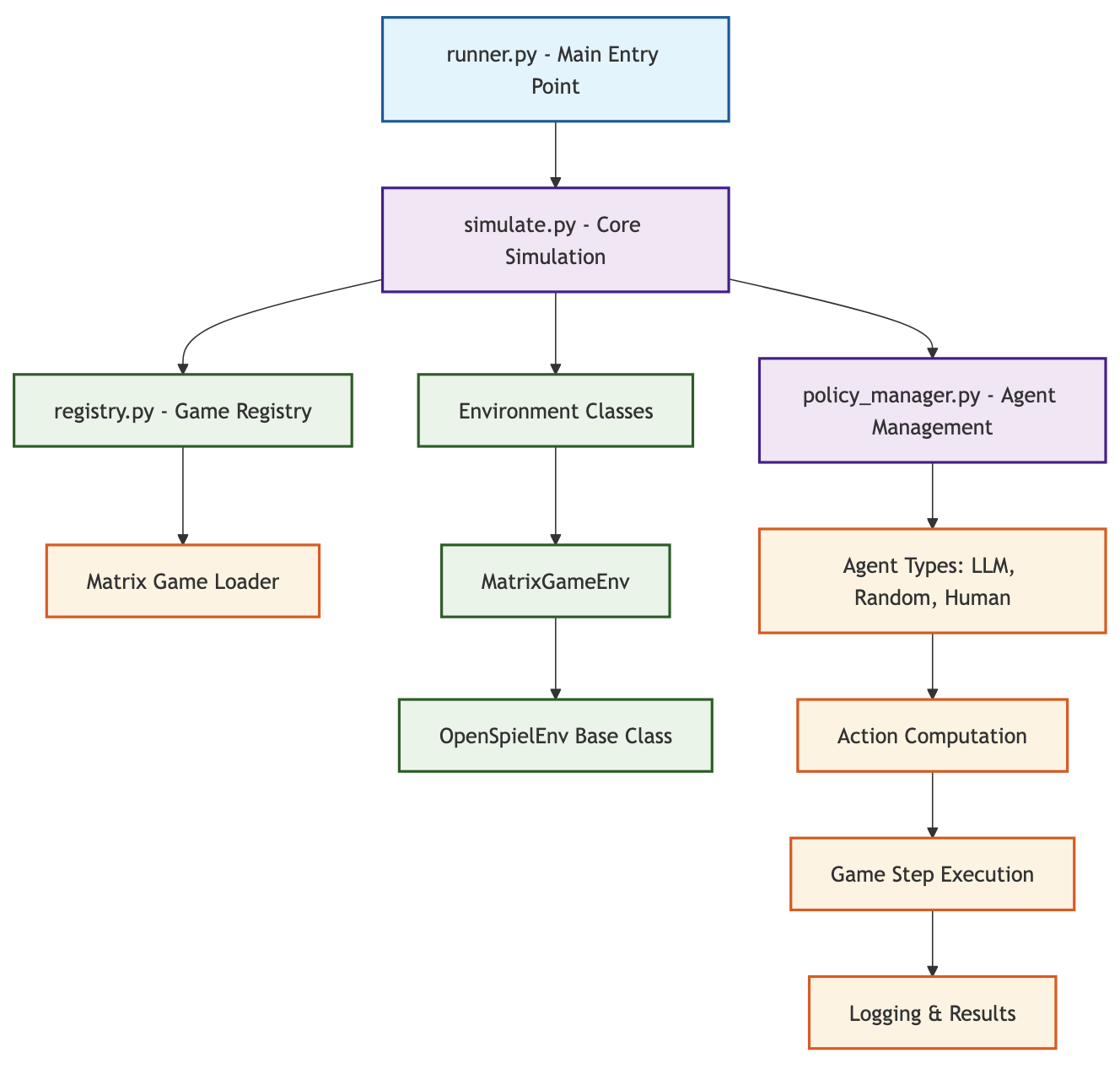}
  \caption{High-level architecture of the Game Reasoning Arena framework for matrix games.}
  \label{fig:llm_arena_architecture}
\end{figure}

The architecture of our codebase is organized into modular components that separate concerns across environment design, agent control, and simulation execution. As illustrated in Figure~\ref{fig:llm_arena_architecture}, the simulation process begins with \texttt{runner.py}, which serves as the main entry point and initializes the overall configuration. This script delegates the core simulation logic to \texttt{simulate.py}, which orchestrates the interaction between the environment, game registry, and agent policies. On the left-hand side of the diagram, \texttt{simulate.py} interfaces with \texttt{registry.py}, a module responsible for loading and registering matrix games through the \textit{Matrix Game Loader}. At the center of the architecture, environment classes such as \texttt{MatrixGameEnv} are defined to implement the game logic and extend a common \texttt{OpenSpielEnv} base class to ensure compatibility with the OpenSpiel interface. On the right-hand side, agent configuration and management are handled by \texttt{policy\_manager.py}, which supports multiple agent types, including large language models (LLMs), random agents, and human-controlled agents. Each agent computes its actions based on the current game state and proceeds through the simulation loop via the action computation and game step execution phases. Finally, game outcomes and agent behaviors are recorded through a structured logging mechanism, enabling detailed analysis and post-hoc evaluation. This modular and extensible design facilitates efficient benchmarking of LLM strategies in matrix game settings.

\begin{figure}[h]
  \centering
  \includegraphics[width=0.99\textwidth]{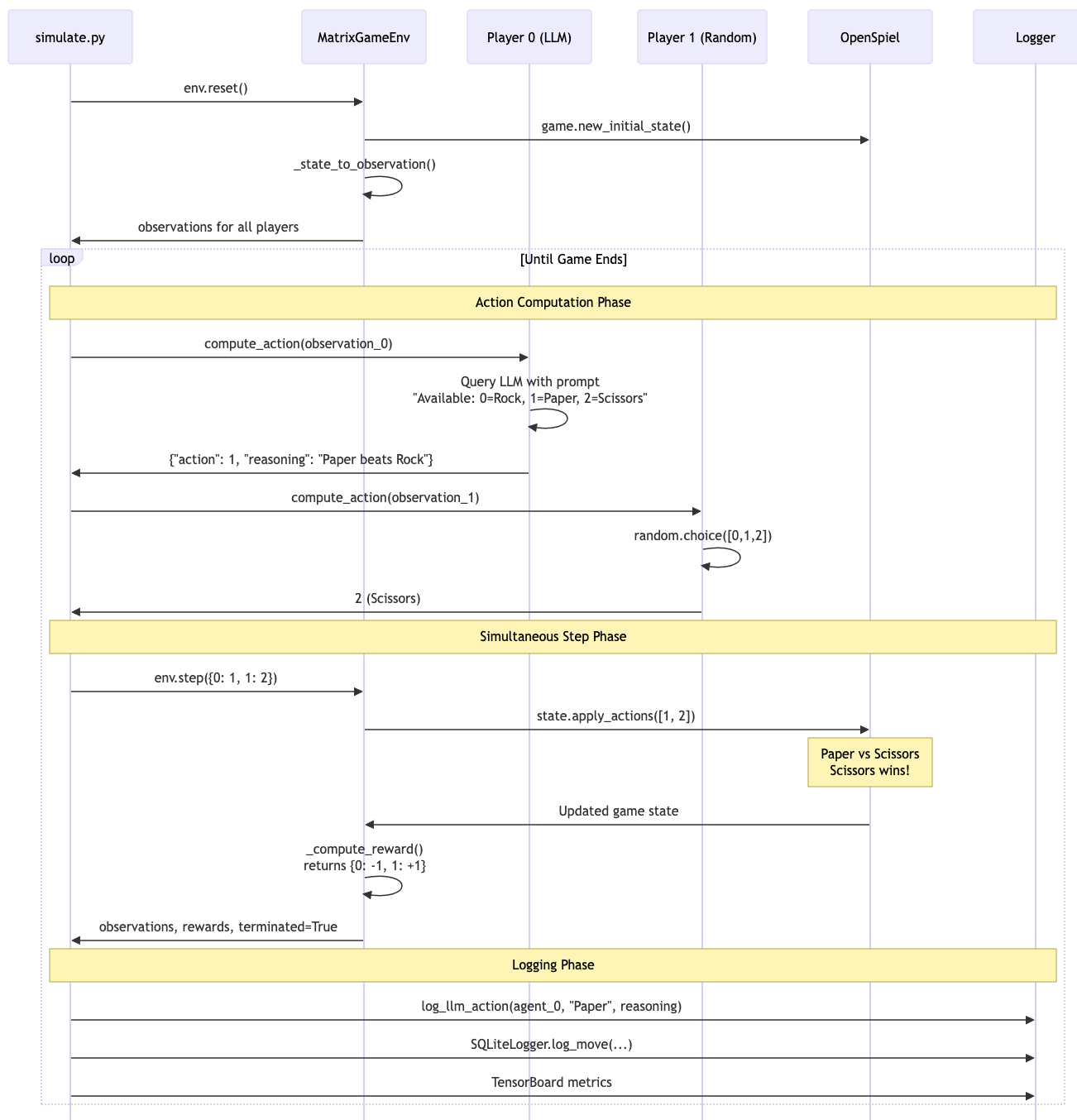}
  \caption{Sequence diagram of a single matrix game episode in the Game Reasoning Arena. It illustrates environment setup, LLM and random agent decision-making, game state updates, and structured logging.}
  \label{fig:code_flow}
\end{figure}

The interaction flow of a single matrix game episode within the Game Reasoning Arena is depicted in Figure~\ref{fig:code_flow}. The simulation begins with environment initialization via \texttt{env.reset()}, followed by the construction of player-specific observations using the \texttt{\_state\_to\_observation()} method. The simulation loop then enters the \textit{Action Computation Phase}, where each agent selects an action based on its observation. For Player 0 (an LLM-controlled agent), this involves querying a language model using a prompt that includes the legal actions. The LLM returns both an action and its reasoning (e.g., ``Paper beats Rock''). For Player 1 (a random agent), actions are sampled uniformly. In the \textit{Simultaneous Step Phase}, the environment executes the joint action using \texttt{env.step()}, which internally calls \texttt{state.apply\_actions()} to apply the moves and computes rewards via \texttt{\_compute\_reward()}. After each step, the environment returns updated observations, rewards, and a termination flag. Finally, during the \textit{Logging Phase}, the system records all relevant information, including the LLM's reasoning, action choices, and outcome metrics, using structured logs (e.g., \texttt{SQLiteLogger}) and real-time visualization tools such as TensorBoard. This loop continues until the game reaches a terminal state.

\end{document}